# Development of knowledge Base Expert System for Natural treatment of Diabetes disease

Sanjeev Kumar Jha
University Department of Mathematics,
Babasaheb Bhimrao
Ambedkar Bihar University, Muzaffarpur - 842001
Bihar (India)

D.K.Singh
University Department of Mathematics,
Babasaheb Bhimrao
Ambedkar Bihar University, Muzaffarpur - 842001
Bihar (India)

*Abstract*—The development of expert system for treatment of Diabetes disease by using natural methods is new information technology derived from Artificial Intelligent research using ESTA (Expert System Text Animation) System. The proposed expert system contains knowledge about various methods of natural treatment methods (Massage, Herbal/Proper Nutrition, Acupuncture, Gems) for Diabetes diseases of Human Beings.
The system is developed in the ESTA (Expert System shell for Text Animation) which is Visual Prolog 7.3 Application. The knowledge for the said system will be acquired from domain experts, texts and other related sources.

## I. INTRODUCTION

This article presents the conceptual framework of natural treatment methods available for diabetes. The main goal of this research is to integrate all the natural treatment information of diabetes in one place.

Expert System named as Sanjeevani is developed using ESTA (Expert System Shell for Text Animation) as knowledge based system to describe the various Natural therapy methods for treatment of Diabetes disease and various other diseases.

The main purpose of the present study is in the design and development of an expert system which provides the information of different types of natural treatment (Massage, Acupuncture, Herbal/Proper Nutrition and gems) of Diabetes. The system background starts with the collection of information of different methods of treatment available for Diabetes diseases. The acquired knowledge is represented to develop expert System.

## II. DEVELOPMENT OF EXPERT SYSTEM

We are in the process of development of Sanjeevani Natural therapy expert system that is developed in ESTA Application. It is designed for assisting in treatment of the Diabetes disease and various other diseases which can be cure naturally with different methods (Massage, Herbal/Proper Nutrition, Acupuncture, Gems) and provide treatment solutions.

The natural therapy consists of a variety of natural body therapies, soul therapies and energy therapies for healing, that we found helpful for peoples health and wellness and which don't cost the earth

The natural treatment is the process of healing/curing diseases through natural, drugless and the most harmless process. Human beings are an intrinsic part of the nature. Therefore, nothing except nature can facilitate a complete cure for human machinery disorders.

These are the various methods of Natural therapy:--

- Massage
- Acupuncture
- Herbal/Proper Nutrition
- Gems

| Diabetes Diseases |
|---|
| Treatment Methods |
| 1) Natural Care(Herbal / Proper Nutrition)<br>2) Acupuncture<br>3) Homeopathic<br>4) Massage<br>5) Gems |
| Treatment Solutions Advice |

Figure 1: Representation of Natural Treatment of Diabetes Diseases

## III. DESIGN OF EXPERT SYSTEM

The Expert System can be created by using ESTA by building the knowledge base.

Expert System --→ ESTA + Knowledge Base

The Quality of Expert System is depends on its knowledge base.

The process of developing knowledge base is:-

a) Identifying the input of Problem
b) Gaining Knowledge
c) Representation of Knowledge

### A. Identifying the Input of Problem

For developing the expert system first we have to identify the problem and its behaviors.

The Input for our system is regarding identifying the different types of natural treatments (Massage, Acupuncture,

Herbal/Proper Nutrition and gems) available for Diabetes Disease.

### B. Gaining Knowledge

Gaining Knowledge is very important in developing the expert system. The acquired gained knowledge is analyzed and processed to give the best solution of the problem.

The knowledge of different types of Natural treatments (Massage, Acupuncture, Herbal/Proper Nutrition and gems) of Diabetes has been gained by reading books, browsing Internet and also got information from consultation of Physician in their respective areas.

### C. Representation of Knowledge

Representation of knowledge is the last phase of the development of knowledge base system. There are various approaches for representation of Knowledge into knowledge base.

Each Knowledge base contains rules for a specific domain. Thus, for a natural treatment of diabetes expert system the knowledge base will contain rules relating certain natural treatment methods of diabetes such as Massage, Acupuncture, Herbal/Proper Nutrition and gems.

ESTA has all facilities to write the rules that will make up a knowledge base. Further, ESTA has an inference engine which can use the rules in the knowledge base to determine which advice is to be given to the expert system user or to initiate other actions

Representation in ESTA is the rule based in logical paradigm of simple if-then rules in backward or forward chaining. We have chosen here the backward chaining for knowledge representation with simple if-do pair in place of if-then rules. Here we have considered two major knowledge representations namely Sections and Parameters. The top level of representation of knowledge in ESTA is section. It contains the logical rules that direct the expert system how to solve problem, actions to perform such as giving advice, going to other sections, calling to routines etc. The first section in ESTA is always named as start section. The advice is given when condition(s) in the section is (are) fulfilled. Parameters are used as variable and it determines the flow of control among the sections in the Knowledge Base. A parameter can be one of the four types: Boolean or logical, Text, Number and Category parameters.

We have developed the various Parameters and Sections for developing this expert system.

## IV. REPRESENTATION OF SANJEEVANI EXPERT SYSTEM

The Knowledge representation in ESTA is based on the items: a) Section b) Parameters c) Title

### A. Representation of Parameters Used in Developing Expert System

In FIGURE 1: Here we have defined the disease parameter which is of type category describing the various types of disease.

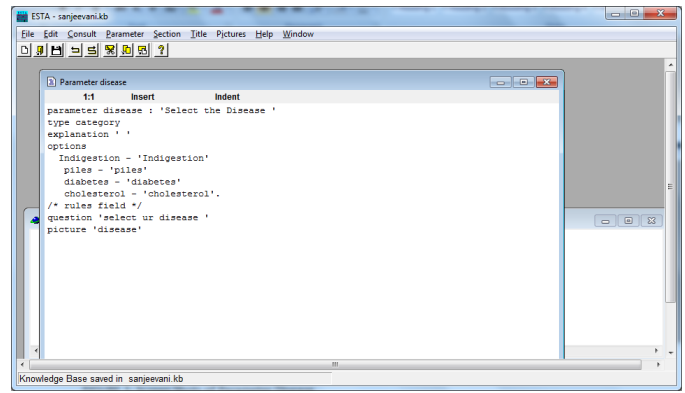

Figure 1. Screen Shots of Parameter Disease

In FIGURE 2: Here we have defined the diabetesop parameter which is of type category describing the various types of natural treatment available for diabetes disease.

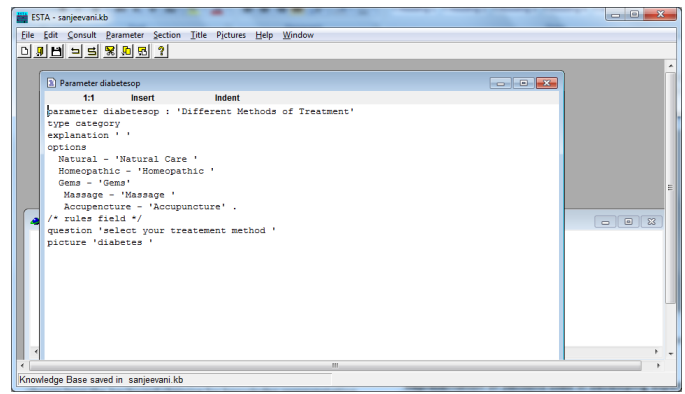

Figure 2. Screen Prints of Parameter diabetesop

### B. Representation of Sections used in developing expert system

In FIGURE 3: Here we have a main section start which is developed to transferring controls in accordance with the user's response about disease

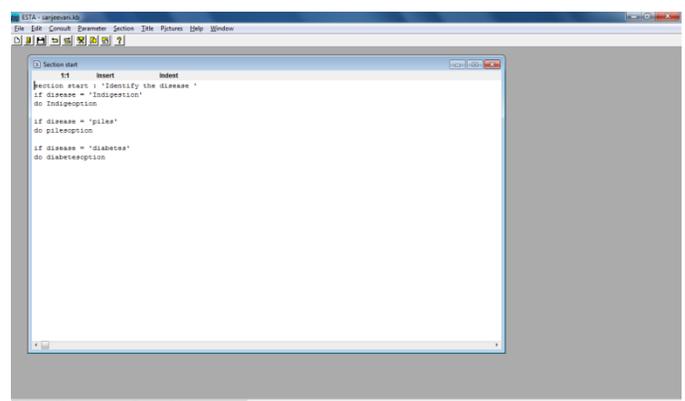

Figure 3. Screen Print of Start Section

In FIGURE 4: Here causeofdiabetes Section describes the diabetes disease and its symptoms

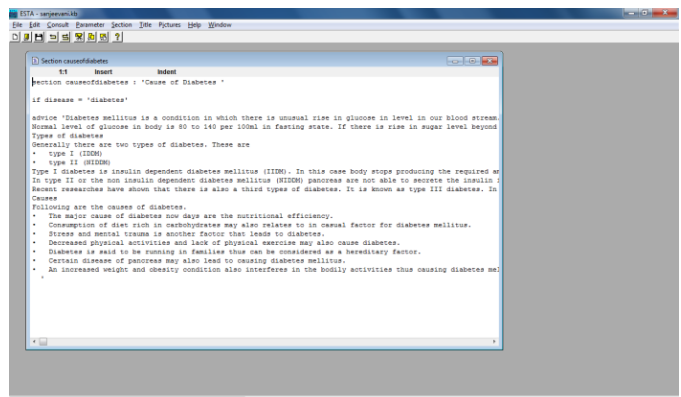

Figure 4. Screen Print of Causeofdiabetes Section

In FIGURE 5: Here in the diabetesoption section describes the various natural treatment options available for diabetes disease.

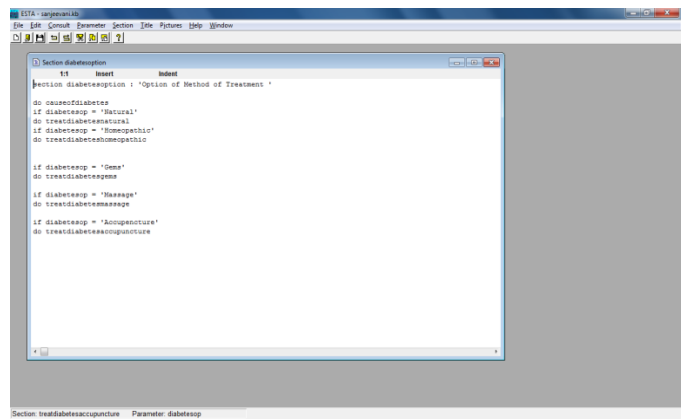

Figure 5. Screen Print of Section of diabetesoption

In FIGURE 6: Here in the treatdiabetesnatural section describes the Natural Care (Herbal / Proper Nutrition) treatment solution of diabetes disease

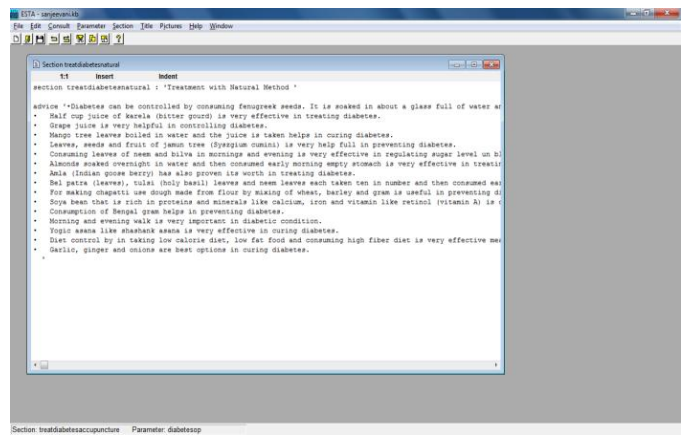

Figure 6. Screen Print of Section treatdiabetesnatural

### 1.1. Representation of Title

In FIGURE 7: It describes the Title of Sanjeevani expert system.

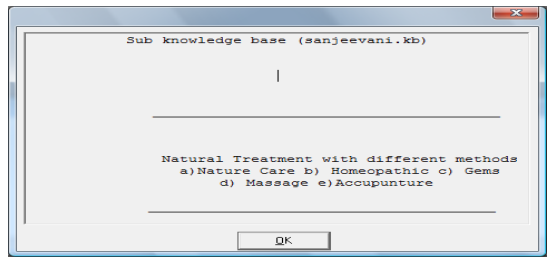

Figure 7. Screen Print of Title of expert System

### V. CONSULTATION OF EXPERT SYSTEM

In FIGURE 8: It describes the beginning of Consultation of Sanjeevani Expert System. It will ask the users to select the disease (Diabetes) for which they want different type of natural treatment solution.

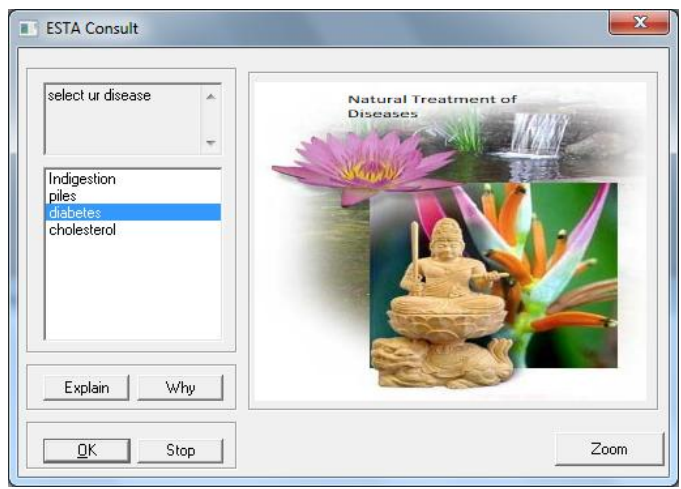

Figure 8. Screen Print of ESTA Consult of Diabetes

In FIGURE 9: It describes the diabetes diseases and its symptoms

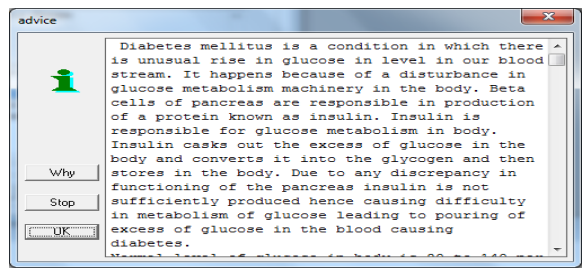

Figure 9. Screen Print of Diabetes description and Symptoms

In FIGURE 10: It describes the different types of Natural treatment methods available for Diabetes disease. It will ask users to select one of the Natural treatment methods for getting details treatment advice.

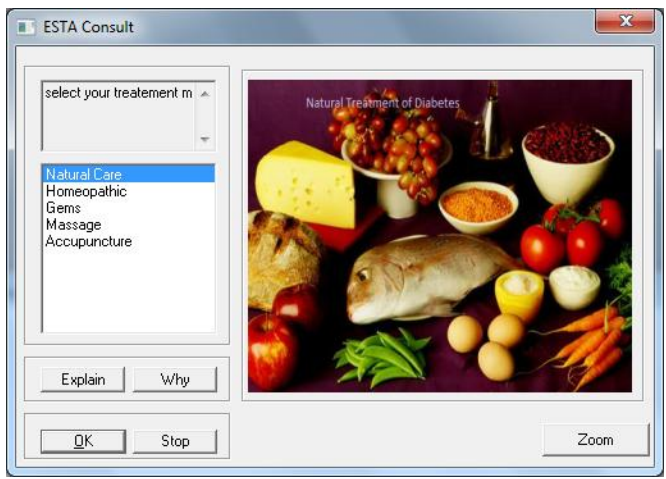

Figure 10. Screen Print of selection of Natural Treatment Method

In FIGURE 11: It describes the Natural Care (Herbal / Proper Nutrition) treatment solution of diabetes disease

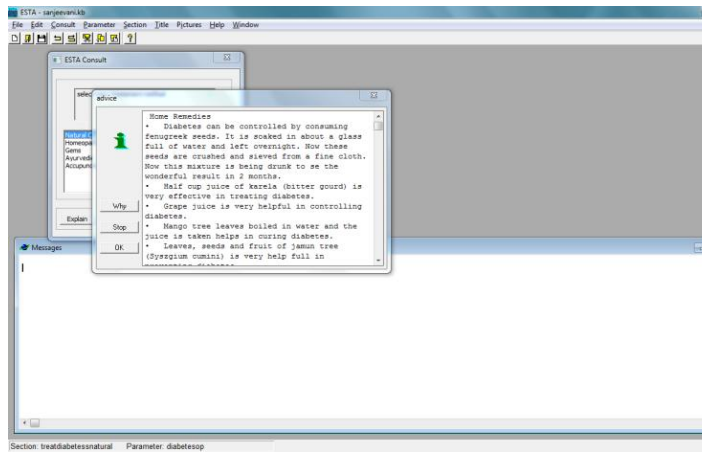

Figure 11. Screen Print of Natural Treatment Solution

## VI. CONCLUSIONS

As the Sanjeevani project shows, applied work in developing of expert system describing the various Natural therapy methods for treatment of Diabetes disease and various other diseases.

The field of medical artificial intelligence is particularly appealing to the physicians and computer scientists working in the area.

The long-term challenges are well recognized-such as the need for mechanisms that will assure completeness and shared knowledge bases for different natural methods treatment of different diseases.

This Expert System development is one of the steps to get the integrated knowledge base system which will help in getting information of various natural treatment methods available for disease to general users (Patient and Physician)

AUTHOR'S PROFILE

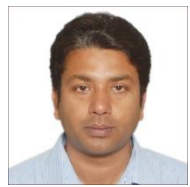

Sanjeev Kumar Jha has received M.C.A from Madras University in 2002; Presently He is pursuing PhD from Ambedkar Bihar University, Muzaffarpur. He has a Software Industry experience of more than 7 years. His areas of interest include Artificial Intelligence, Expert System and Software Testing.